\documentclass[preprint,12pt]{elsarticle}

\usepackage{amssymb}
\usepackage{amsmath}

\usepackage{booktabs}
\usepackage{csquotes}
\usepackage{listings}
\journal{Nuclear Physics B}

\begin{document}

\begin{frontmatter}



\title{APD-Agents: A Large Language Model-Driven Multi-Agents Collaborative Framework for Automated Page Design}


\author[label1]{Xinpeng~Chen}
\author[label3]{Xiaofeng~Han}
\author[label2]{Kaihao~Zhang}
\author[label3]{Guochao~Ren}
\author[label3]{Yujie~Wang}
\author[label3]{Wenhao~Cao}
\author[label3]{Yang~Zhou}
\author[label1]{Jianfeng~Lu}
\author[label1]{Zhenbo~Song\corref{cor1}}

\cortext[cor1]{Corresponding author. E-mail: songzb@njust.edu.cn}

\affiliation[label1]{
    organization={School of Computer Science and Engineering, Nanjing University of Science and Technology},
    city={Nanjing},
    country={China}
}

\affiliation[label2]{
  organization={School of Computer Science and Technology, Harbin Institute of Technology},
  city={Shenzhen},
  country={China}
}

\affiliation[label3]{
  organization={Huatai Securities Co., Ltd.},
  city={Nanjing},
  country={China}
}

\begin{abstract}
Layout design is a crucial step in developing mobile app pages. However, crafting satisfactory designs is time-intensive for designers: they need to consider which controls and content to present on the page, and then repeatedly adjust their size, position, and style for better aesthetics and structure. Although many design software can now help to perform these repetitive tasks, extensive training is needed to use them effectively. Moreover, collaborative design across app pages demands extra time to align standards and ensure consistent styling. In this work, we propose APD-agents, a large language model (LLM) driven multi-agent framework for automated page design in mobile applications. Our framework contains OrchestratorAgent, SemanticParserAgent, PrimaryLayoutAgent, TemplateRetrievalAgent, and RecursiveComponentAgent. Upon receiving the user's description of the page, the OrchestratorAgent can dynamically can direct other agents to accomplish users' design task. To be specific, the SemanticParserAgent is responsible for converting users' descriptions of page content into structured data. The PrimaryLayoutAgent can generate an initial coarse-grained layout of this page. The TemplateRetrievalAgent can fetch semantically relevant few-shot examples and enhance the quality of layout generation. Besides, a RecursiveComponentAgent can be used to decide how to recursively generate all the fine-grained sub-elements it contains for each element in the layout. Our work fully leverages the automatic collaboration capabilities of large-model-driven multi-agent systems. Experimental results on the RICO dataset show that our APD-agents achieve state-of-the-art performance.
\end{abstract}




\end{frontmatter}



\section{Introduction}
\label{intro}
Layout design is a pivotal stage in mobile application development, as an intuitive and consistent interface greatly enhances user experience. However, being proficient in using a design software requires significant training. And manual coordination among designers to maintain stylistic consistency across pages can also cost lots of time. 

\begin{figure}[htbp]
\centering
\includegraphics[width=\linewidth]{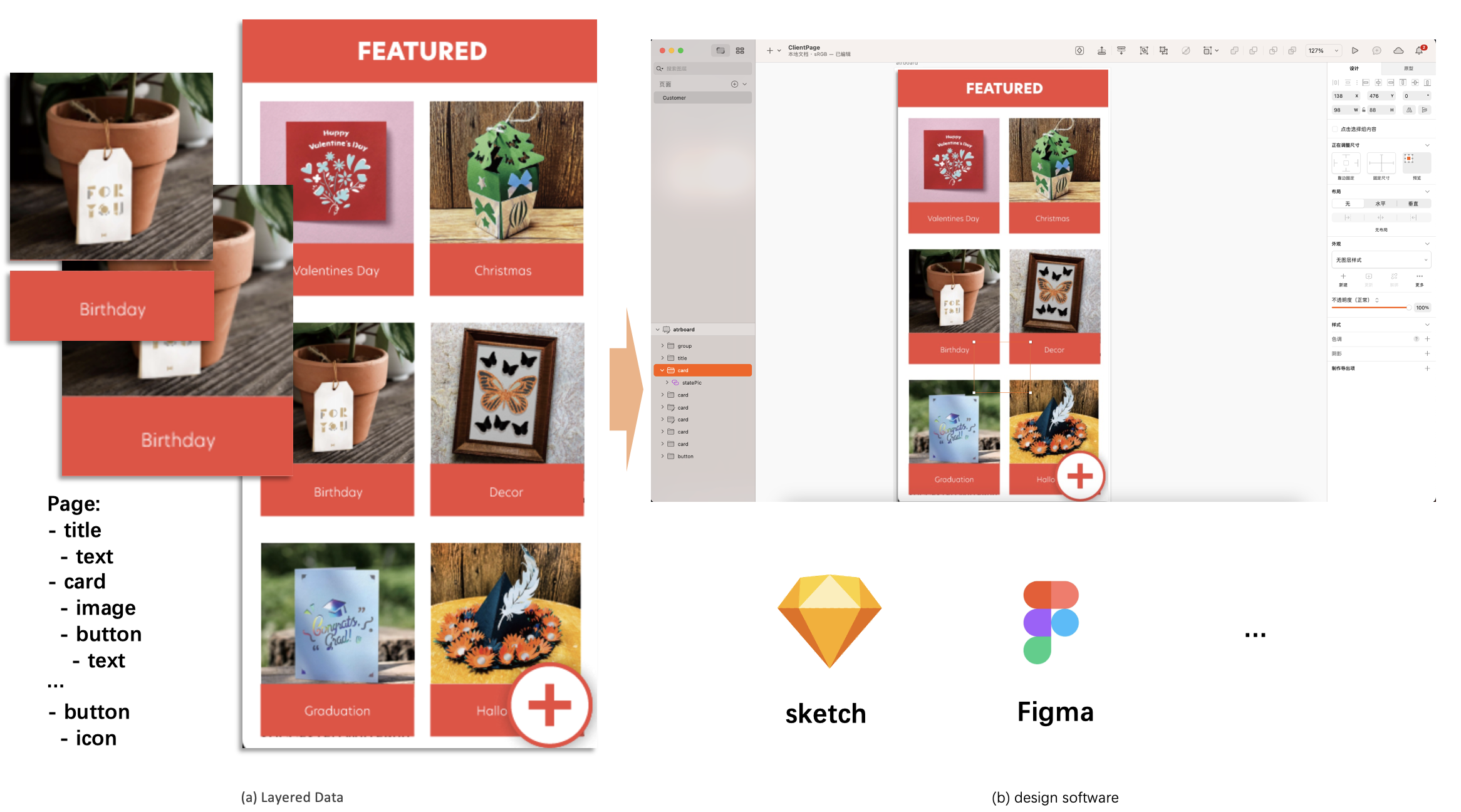}
\caption{An example of page design with Sketch. The elements of one page should be layered so that the designers can modify them via Sketch, Figma or other design software.}
\label{sketch}
\end{figure}

To alleviate these burdens, recent research has explored AI-driven solutions for layout generation and refinement~\cite{fujitake2024layoutllm,gupta2021layouttransformer,li2020layoutgan,patil2020read,lee2020neural,rahman2021ruite}. These approaches aim to reduce the time and expertise required for high-quality cohesive designs by automating element selection, arrangement, and styling. Most of these methods rely on computer vision deep neural networks that, given descriptions of target page elements, automatically generate layout results, typically as static images. However, such outputs cannot be edited in design software like Sketch or Figma, where design data is organized as hierarchical layers and must adhere to a fixed structured format (see Fig.~\ref{sketch}). Moreover, certain approaches~\cite{kong2022blt,lee2020neural,li2020attribute} are tailored to specific tasks or user constraints, greatly limiting both the diversity of user requirements and the extensibility of layout generation. In addition, most existing techniques~\cite{gupta2021layouttransformer,li2020attribute,li2020layoutgan,jyothi2019layoutvae} struggle to handle complex pages involving many elements and a wide variety of component types, and thus cannot effectively support designers in realistic work scenarios. Beyond these obstacles, all of these methods depend on large collections of annotated design drafts for model training; given the significant variation in design specifications across different websites and applications, a model trained on one dataset often generalizes poorly to another, and the cost of acquiring annotated data for each new specification is prohibitively high, making it nearly impossible to train a single unified network model.

Fortunately, since 2023 large language models (LLMs) have demonstrated impressive performance on various tasks~\cite{achiam2023gpt,nori2023capabilities,wei2022chain,wei2022emergent} and increasingly excel at generating structured data~\cite{jiang2023structgpt}. In this paper, we propose APD-agents, a large language model driven multi-agents collaborative framework for automated page design. Our method can sequentially design the page in a coarse‑to‑fine way, and output the structured JSON data that can be converted into data formats recognizable by software such as Sketch and Figma. Specifically, APD-agents contain several agents as follows: an OrchestratorAgent which can plan the task and manage other agents, a SemanticParserAgent to extract elements, a TemplateRetrievalAgent and a PrimaryLayoutAgent to generate the initial layout, and a RecursiveComponentAgent to generate sub layouts recursively. 

Our main contributions are as follows:

1. We propose a hierarchical generation multi-agents framework, in which the OrchestratorAgent schedules specialized agents, including PrimaryLayoutAgent for coarse layouts, RecursiveComponentAgent for recursive refinement, and TemplateRetrievalAgent for retrieve few‑shot examples to enhance the design performance of complex pages without training.

2. Our output is structured JSON data rather than images, ensuring that it can be easily converted into data formats loadable by professional design software.

3. Empirical validation on the RICO dataset \cite{deka2017rico} demonstrates state‑of‑the‑art performance of our method.

\section{Related Works}
\subsection{LLM-driven multi-agent system}
Recent work has shown that integrating large language models into multi-agent systems enables advanced reasoning, planning and collaboration across diverse tasks by sharing a schema-based knowledge store and leveraging agent specialization for modularity, parallel execution and fault isolation \cite{forouzandehmehr2025cal,yuan2024maxprototyper,xi2025multi}. Frameworks such as AutoGen \cite{autogen} further reduce development overhead by providing abstractions for agent roles, memory management and inter-agent communication. Building on these advances, our APD-agents decomposes the layout generation task into specialized sub-agents that operate independently within a shared context, thereby improving both efficiency and result quality.

\subsection{Automated page design}
There are many works on layout generation in recent years. Layout generation tasks include document layout generation \cite{fu2022doc2ppt,zheng2019content,zhong2019publaynet,he2023diffusion,yamaguchi2021canvasvae}, user interface layout generation \cite{kong2022blt,deka2017rico,lin2023parse}, etc. In this work, we mainly focus on the interactive interface layout generation tasks when users use mobile apps. Layout generation can be mainly classified into 4 different types: compute vision based methods, natural language processing based methods, multi-modal based methods and multi-agent based models.

\subsubsection{Compute Vision based methods}
These methods focus mainly on leveraging visual information to generate effective layouts. For example, LayoutGAN \cite{li2019layoutgan} utilizes a Generative Adversarial Network (GAN) framework to model the geometric relationships among various types of 2D elements, optimizing the generated layouts using a CNN-based discriminator. LayoutVAE \cite{jyothi2019layoutvae} is a variational auto encoder-based framework that generates complete image layout given a set of element labels. LayoutTransformer \cite{gupta2021layouttransformer} leverages a self-attention mechanism \cite{vaswani2017attention} to learn the contextual relationships between layout elements and generate new image layouts within a specified domain. LayoutDiffusion \cite{zhang2023layoutdiffusion} introduces a novel generative model for automatic layout generation,utilizing a block transfer matrix combined with a piecewise linear noise table to unify the classification and ordinal properties of the layout.

\subsubsection{Natural Language Processing based methods}
Models based on visual information typically require large-scale datasets for training and are often limited to handling single tasks based on specific inputs, such as element type and size. These constraints pose challenges in meeting the diverse needs of users in practical applications. To overcome  these limitations, natural language processing based models have been developed, leveraging advancements in large-scale models to analyze users' generative requirements expressed in natural language and produce corresponding layouts. For instance, previous method \cite{lu2023ui} uses UI grammar to represent the hierarchical relationships between UI elements and guides large models in generating layouts through context learning. Menucraft \cite{kargaran2023menucraft} generates textual layout suggestions by parsing natural language descriptions of user intentions and goals using large models. Similarly, approaches like Layout Prompter \cite{lin2024layoutprompter}, utilize prompt strategies within large models to generate layouts effectively.

\subsubsection{Multi-modal based methods}
With the increasing generalization capabilities of large language models (LLMs), recent studies have explored applying layout generation tasks to multi-modal models. For instance LayoutGPT \cite{feng2024layoutgpt}, which leverages large models to convert language inputs, including spatial relationships and numerical data, into text-to-image layout generation. This approach has also demonstrated promising results in the synthesis of 3D interior scenes. Similarly, LayoutLLM \cite{fujitake2024layoutllm} uses multi-modal large models to interpret   document information within layouts, introducing a layout instruction adjustment strategy to improve layout comprehension. PosterLLaVa \cite{yang2024posterllava} employs structured text (JSON format) and visual instruction tuning to generate layouts under specific visual and textual constraints, including user-defined natural language specifications.

\begin{figure}[!t]
\centering
\includegraphics[width=\linewidth]{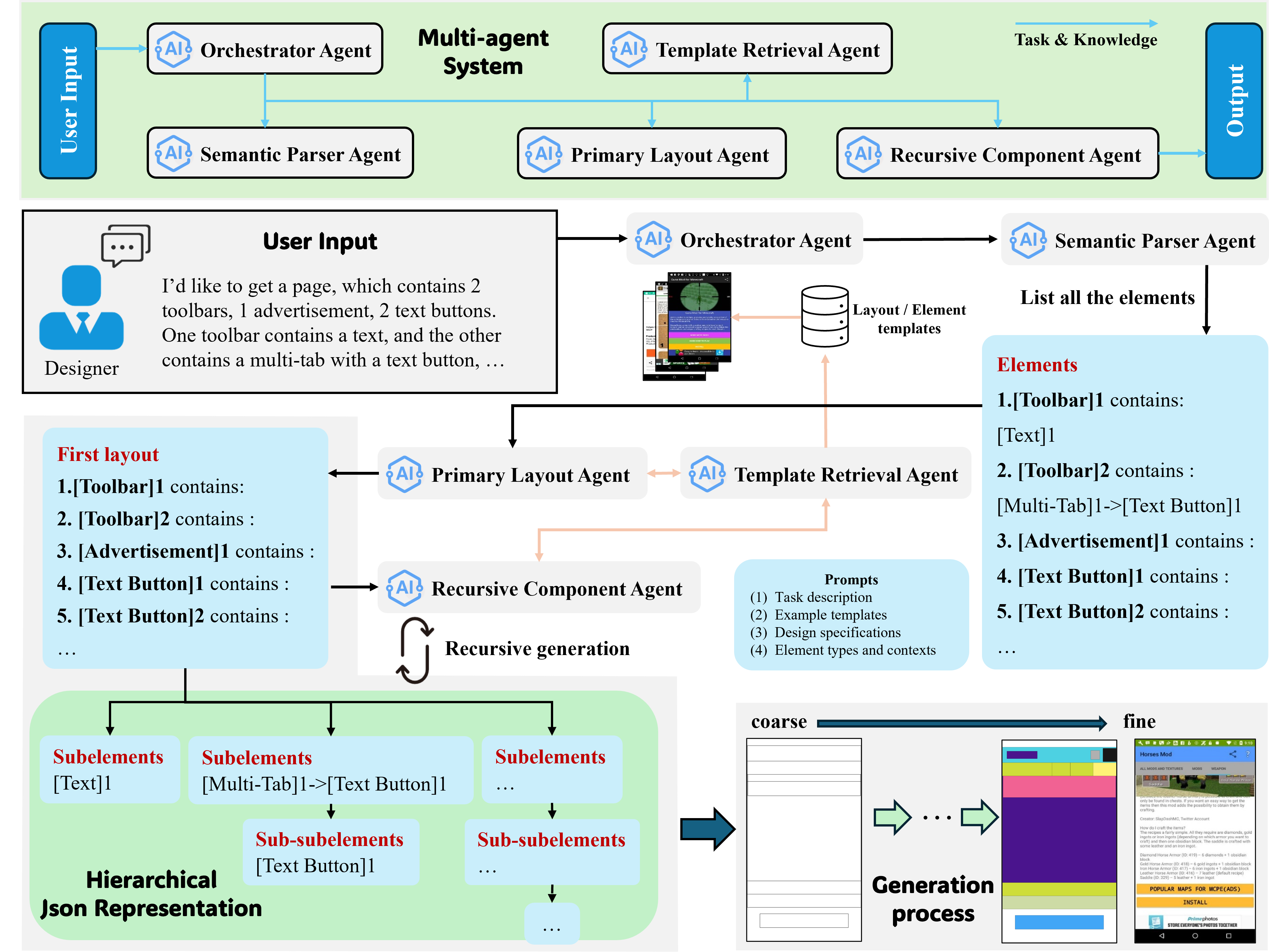}
\caption{Our APD-agents framework.}
\label{Figureworkflow}
\end{figure}

\section{Method}
In this section, we first describe the task, and then present the framework and details of our method.

\subsection{Task Formulation}
In a layout generation task, designers describe page elements and functionality in natural language. An LLM interprets these specifications to list element types and contents, after which we determine each element’s bounding-box (position and size) and style (color, appearance, and related attributes).

\begin{figure}[htbp]
  \centering
  \includegraphics[width=\linewidth]{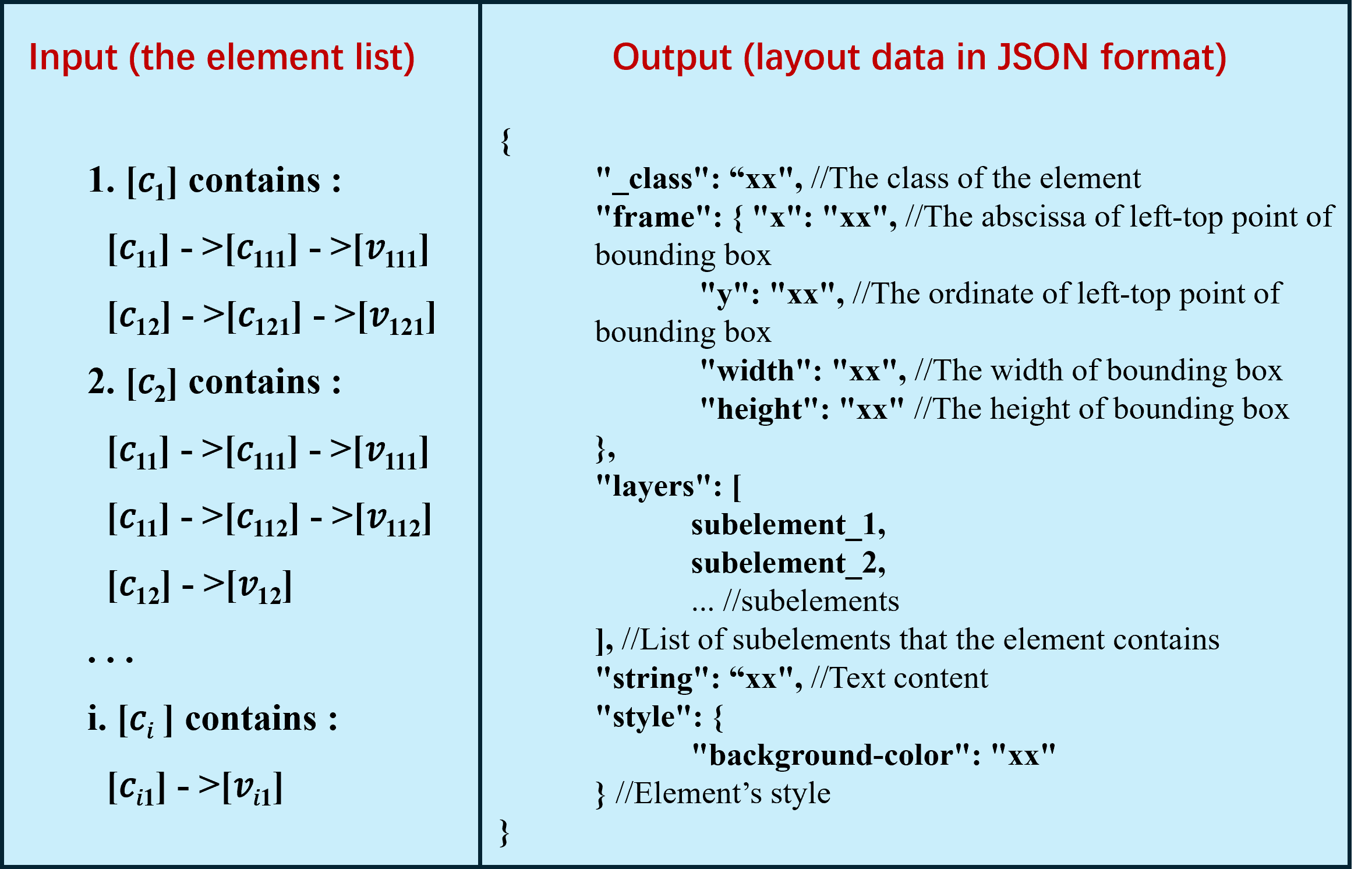}
  \caption{Example of the input data and output schema. In the output format, the sub elements in key \enquote{layers} have the same json format as the output.}
  \label{input_output}
\end{figure}

As the input data and output schema shown in Figure \ref{input_output}. Let's define all the elements of a page as $E$ and the number of them as $M$. Each element $e_i$ has following properties: the class property $c_i$, the geometric properties $x_i$, $y_i$, $w_i$, $h_i$, and the value property $v_i$. Meanwhile, we adopt the \enquote{\texttt{->}} notation to denote the mapping between subelements and their parent element or associated value. For instance, \enquote{$a\texttt{->}b$} indicates that $b$ is a subelement or value of $a$.
Here $c_i$ represents the element category label, such as text, image, button and so on. As we have defined the elements' hierarchy, we use $c_{ij}$ to represent the class label of j-th subelement $e_i$. $x_i$ and $y_i$ are the coordinates of the left-top point of its bounding box, and $w_i$ and $h_i$ are the width and height. $v_i$ is the value of $e_i$. For a text element, $v_i$ represents the character content it displays, while for a image or a icon element, $v_i$ is the name. Besides, since we need to retrieve template pages and elements as references, we note the number of all the existing pages and elements as $N_{p}$ and $N_{e}$. The final output data of the target page is $p_i$, which is a text data that conforms to the JSON format definition.

\subsection{Architecture Overview}
The framework of our method is shown in Figure \ref{Figureworkflow}. 

In our multi-agent framework, the user’s natural language description is first submitted to the OrchestratorAgent, which coordinates all subsequent subtasks. The SemanticParserAgent converts the input into a hierarchical list of elements; the PrimaryLayoutAgent retrieves first-layer layout exemplars from the TemplateRetrievalAgent and produces a coarse initial arrangement; finally, the RecursiveComponentAgent examines the current JSON structure and recursively refines subcomponents. This top-down, iterative generation process combines element data, exemplar templates, and design specifications as prompts to yield a complete page layout (see Figure \ref{Figureworkflow}).

\subsection{Agent Roles and Responsibilities}
\subsubsection{Orchestrator Agent}
In our system, the OrchestratorAgent serves as the entry point and task scheduler. It ingests the user’s natural‑language specification, interprets the task intent, and dispatches subtasks to specialized agents according to predefined policies. Additionally, it continuously monitors the progress and intermediate outputs of each subtask, aggregating them into the final hierarchical JSON layout once all subtasks have completed. The system message OrchestratorAgent used can be found in the appendix.

\subsubsection{Semantic Parser Agent}
The SemanticParserAgent transforms the user’s natural‑language description into a structured symbolic representation. Specifically, using tailored prompts, this agent extracts the required page elements and their attributes, infers their hierarchical relationships, and reconstructs the initial element tree, thereby providing a standardized input for subsequent template retrieval and layout generation.

\subsubsection{Template Retrieval Agent}
One of the most significant capabilities of large language models is In‑Context Learning (ICL): by incorporating a few similar exemplars, the model can follow their input–output patterns without additional training, yielding more accurate and high‑quality outputs. Accordingly, the TemplateRetrievalAgent provides high‑quality few‑shot exemplar retrieval during layout generation. 

In our approach, we first collect the existing design pages with number $N_p$, and reorganize them into the standard input format. We save all the elements according to their types $c_i$. In the saved data, each element template contains not only its attributes $x_i$, $y_i$, $w_i$, $h_i$, $v_i$, but also its hierarchical path, for example, $[c_1]\texttt{->}[c_{11}]\texttt{->}[c_{111}]$. The attributes and hierarchical path will later be used for retrieval. The layout result of each element is also saved. 

\begin{figure}[htbp]
\centering
\includegraphics[width=\linewidth]{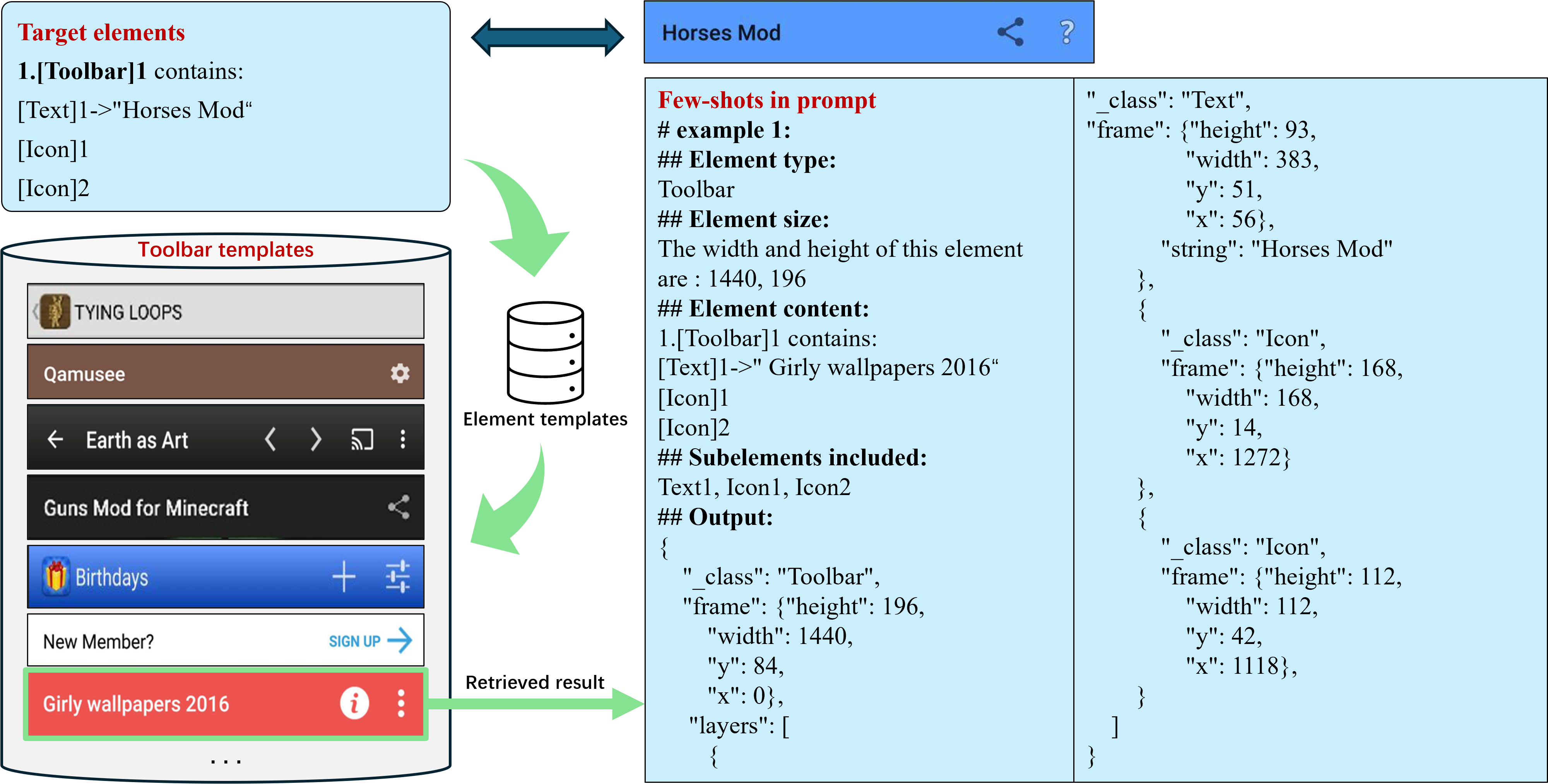}
\caption{Example of template retrieval process. For toolbar to be generated, we retrieve from the element template library and select the most similar one (framed in green in the figure) as the element template to add in the prompt.}
\label{few-shot}
\end{figure}

As we mentioned above, due to the adoption of a coarse-to-fine generation strategy, we have conducted retrieval for elements at different level of the layout generation. In order to utilize the in-context learning capabilities of LLM, these retrieval results are added as few-shot examples in the prompt. As shown in Figure \ref{few-shot}, the element hierarchical path and attributes will be used as the input of the few-shot examples, while the layout result of them will be used as the output of the examples.

In our system, we employ semantic vector encoding retrieval, mapping element descriptions into an embedding space and selecting the most relevant templates based on cosine similarity to serve as prompts. In this kind of retrieval method, we send the path and the value of the elements into a embedding model (GLM embedding v2) \cite{DBLP:conf/acl/DuQLDQY022} to generate their semantic vectors $\{Emb_1, Emb_2, ... ,Emb_N\}$. After that, we use the same embedding model to get $Emb_{new}$, the vector of input data of the target element. We calculate the cosine similarity between the semantic vectors of the new element $Emb_{new}$ and the semantic vectors of all element templates in pairs $Emb_i(i \in N)$, then sort them from largest to smallest, and select the top-k templates as the retrieval results. 

\subsubsection{Primary Layout Agent}
The PrimaryLayoutAgent is responsible for generating the first-layer layout of the page following a coarse-to-fine strategy. After receiving the top-level element list from the SemanticParserAgent, it classifies elements into two categories: \enquote{basic} elements such as Text and Image, and \enquote{composite} elements such as Toolbar and Button.

\begin{figure}[htbp]
\centering
\includegraphics[width=\linewidth]{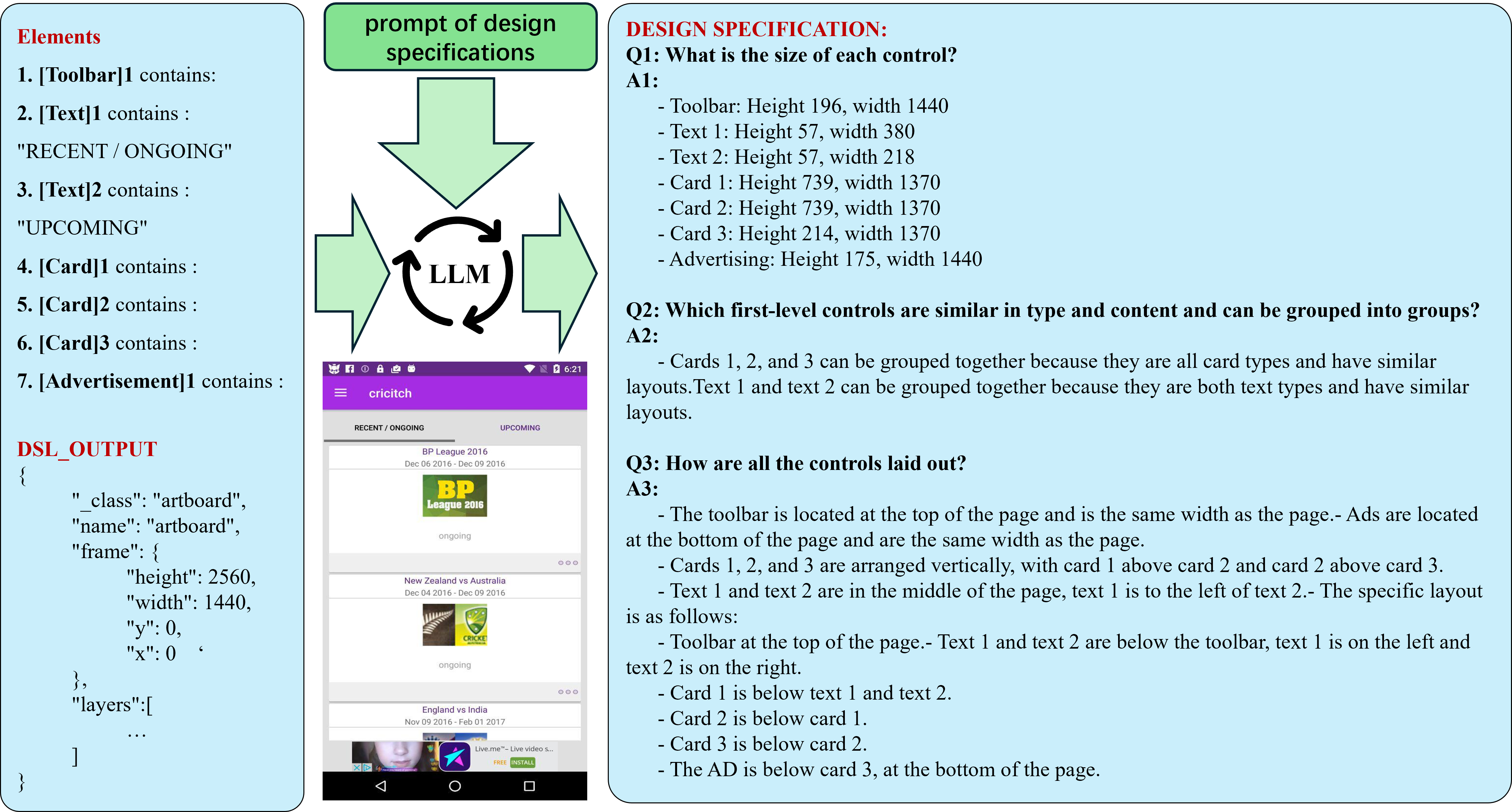}
\caption{Examples of design specification summary. When generating the layout of the first level elements, we send the retrieval results into a LLM to get the design specifications. Prompts for generating design specifications is provided in the appendix.}
\label{FigureDesignSpecification}
\end{figure}

To guide the layout generation, the agent retrieves few-shot exemplars and summarized design specifications from the TemplateRetrievalAgent. The design specifications of pages are reflected in various aspects, such as the style of elements, the color, size, and font of text, the margins of the page, and the spacing between elements. Usually, different mobile applications have different design specifications, and it is challenging to label data that conforms to the specifications for each product and train different models accordingly. Therefore, for all reference elements retrieved, we dynamically summarize the design specifications from them using a large language model, and then we use those specifications to guide the generation of the new page.Figure \ref{FigureDesignSpecification} shows an example of design specification summary via large language model. 

Based on the information above, the PrimaryLayoutAgent uses an LLM to generate the bounding-box properties of each first-layer element, including position, size, and style. The resulting structure serves as the initial layout framework.

\subsubsection{Recursive Component Agent}
The RecursiveComponentAgent refines the initial layout by recursively expanding composite elements. It checks each child node in the first-layer layout against a set of layout criteria, including spatial position, element size, color, and textual content.

If a node does not meet these criteria, the agent extracts its subelements, queries the TemplateRetrievalAgent for relevant few-shot examples, and generates the layout of the next-level components. If the node already satisfies the criteria, it is kept unchanged and recursion stops at that point.

\begin{figure}[htbp]
\centering
\includegraphics[width=\linewidth]{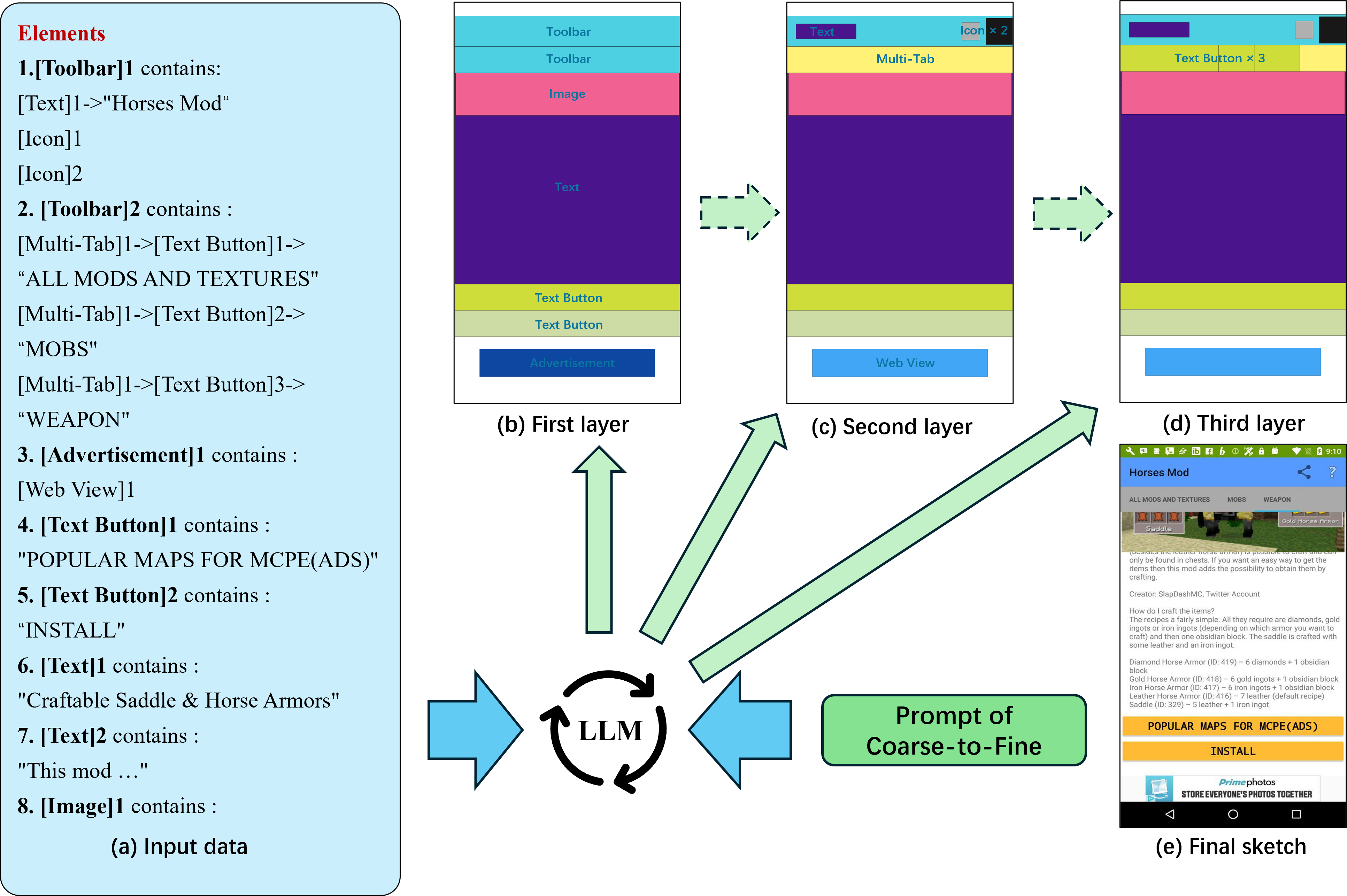}
\caption{The progress of coarse to fine layout generation.}
\label{FigureCoarse2FineGeneration}
\end{figure}

This top-down, depth-first process continues until all elements are resolved into basic components. Once completed, the full layout structure is returned to the OrchestratorAgent for validation and final output. In Figure.~\ref{FigureCoarse2FineGeneration}, (a) shows all the elements contained in this page along with their hierarchy information. From (b) to (d), we start generating the layout from the first-level elements, performing a total of three generations to ensure that all element details are fully generated. Among them, The element name on each layout image represents the newly generated element of this layer relative to the previous layer. (e) gives the representation of the layout in a real APP.

\section{Experiments}

\subsection{Experiment Settings}
\subsubsection{Datasets}
We primarily conduct our experiments using the Rico \cite{deka2017rico} dataset, which contains design data from more than 9.7k Android applications across 27 distinct categories. This dataset offers a comprehensive collection of visual, textual, structural, and interactive design properties, encompassing more than 66k unique user interface screens. All of the layout data are organized according to the schema defined in Figure \ref{input_output}.

Similar to previous works \cite{lin2024layoutprompter}, we randomly split the dataset into two sub-datasets. Of these, 90\% are treated as retrievable reference pages, while the remaining 10\% are used as test data to evaluate layout generation performance. Notably, although most of the data is treated as reference pages, when generating a test page, we retrieve only 500 pages at a time, using them as few-shot examples to dynamically summarize design specifications and guide layout generation. Therefore, in actual design workflows, even with a small number of reference pages, our method can still effectively generate layouts.

\subsubsection{Evaluation Metrics}
To quantitatively evaluate our methodology , we use four evaluation metrics: Maximum Intersection Over Union \cite{kikuchi2021constrained}, Alignment, Overlap \cite{li2020attribute} and Element Properties Accuracy. The explanations of these metrics are as follows: 

Maximum Intersection Over Union (mIoU) measures the overlap between the bounding boxes of a generated element and its corresponding ground truth element. A higher mIoU value indicates better overlap between the generated result and the ground truth.

Alignment (Ali) calculates the minimum distances between the bounding-box edges of element pairs at the top, bottom, left, right, x-center line, and y-center line. It then computes the average of these distances. A smaller value indicates that the elements are more neatly aligned in the generated layout.

Overlap (Ovp) calculates the degree of overlap between all pairs of elements on the same page. A low Ovp value suggests that elements maintain an appropriate distance from each other. For this metric, we only calculate the Ovp between elements of the same class, and then compute the average of these values to obtain the overall Ovp score for the entire generated layout. 

We propose the Element Properties Accuracy (EPAcc) metric to evaluate class, text, style and hierarchy discrepancies between generated pages and their ground truth. Each page is modeled as a tree whose nodes represent elements and whose children represent sub-elements. Our method preserves both tree structure and element order, enabling EPAcc to compare each corresponding node’s class, text and style to compute the overall layout accuracy.

\subsection{Baseline Methods}
We compare our method with the following methods in the experiments.

\textbf{LayoutDiffusion} \cite{zhang2023layoutdiffusion}. This method is a vision-based discrete denoising diffusion model that iteratively refines layouts through mild forward–reverse processes.

\textbf{LayoutFormer++} \cite{jiang2023layoutformer++}. This method employs a constraint serialization scheme to convert user conditions into token sequences, treating layout generation as a sequence-to-sequence task.

\textbf{LayoutPrompter} \cite{lin2024layoutprompter}. This method frames conditional layout tasks for LLMs via input–output serialization, dynamic exemplar selection, and a ranking module. Although similar to our method, its performance is affected by the number of elements in the layout, with a larger number of elements leading to more failure cases.

\textbf{LayoutVAE} \cite{jyothi2019layoutvae}. This method takes the latent code and category labels (optional) as input and generates the element bounding boxes in an autoregressive manner. 

\textbf{LayoutGAN++} \cite{kikuchi2021constrained}.This method improves LayoutGAN
with Transformer backbone and applies several beautifcation post-process for alignment and non-overlap. 

\textbf{NDN} \cite{lee2020neural}.This method is a pipeline system with graph generation, layout synthesis and refinement. 

\textbf{LayoutTransformer} \cite{gupta2021layouttransformer}.This method autoregressively generates a sequence of element tokens.

\textbf{LST-None} \cite{weng2023learn}.This method collaboratively generates layouts by introducing a spatial graph generator and a subsequent layout decoder for knowledge transfer in training and inference.

\textbf{Real Data}. This represents the ground truth of the test data.

\subsection{Implementation Details}
We conducted all baseline evaluations using the original results reported in their respective publications. For our method, we employ Doubao-1.5-pro-32k \cite{guo2025seed1}. We set the sampling temperature to 0.7, enable streaming, and impose a 600s timeout. All other settings remain at its defaults. We build our multi-agent system on the AutoGen framework \cite{autogen}, which simplifies the definition and coordination of multiple interacting agents.

We improve retrieval efficiency by preprocessing and caching all level-1 elements, converting their layout data to conform to our input–output schema before storage. Elements at other levels undergo the same reformatting process shown in Figure \ref{input_output}. Next, we organize elements by class and remove duplicate entries (those sharing both parent path and class) to reduce redundancy and speed up lookups. All prompt templates and system messages are included in the appendix.

\subsection{Main Results}
Previous work \cite{jiang2023layoutformer++} defines six typical layout generation tasks: \textit{Generation conditioned on types} (Gen-T), \textit{Generation conditioned on types and sizes} (Gen-TS), \textit{Generation Conditioned on Types and Sizes} (GEN-TS), \textit{Generation conditioned on relationships} (Gen-R), \textit{Refinement}, \textit{Completion and Unconstrained generation} (UGen). Based on these task definitions, our method aligns most closely with the Gen-T category, so we compare it with several more advanced methods for this task. 

\subsubsection{Quantitative Comparison}
As shown in Table \ref{TableQuantitativeResults}, our method achieves the highest mIoU (0.485), improving by 0.045 over the next best baseline (LST-none, 0.440), which demonstrates more precise element placement. It also attains the lowest overlap error (Ovp = 0.448), a 0.043 reduction compared to LayoutDiffusion (0.491), indicating better coverage without spurious regions. While our alignment error (Ali = 0.192) is slightly above the best (LayoutTransformer, 0.100), it remains on par with most baselines and reflects a balanced trade-off between element recall and spatial coherence.

Additionally, we also report an EPAcc of 85.90\% for our generated pages. We categorize layout mismatches as value mismatches (differences in class, style, or text), class omissions (missing required elements), and key omissions (elements missing one or more attributes). As Table \ref{TableEPAcc} shows, our method includes all required elements, so class omissions and class-related violations are zero. Over 97\% of remaining errors stem from style discrepancies in value and key omissions. We attribute this to two factors: selected reference templates may not cover every element’s style, and in RICO some element styles (icons, text buttons, e.g.) vary with context but are specified to the model only by type, hindering precise style inference.

\begin{table}[t]
\centering
\caption{Specific violation type statistics when calculating EPAcc.}
\begin{tabular}{l|c|c|c|c|c}
    \toprule
    & Class & Style & String & Total & Average \\
    \midrule
    Value & 0 & 32176 & 947 & 33123 & 5.32 \\
    Class & - & - & - & 0 & 0.00  \\
    Key & 0 & 6423 & 136 & 6559 & 1.05 \\
    \bottomrule
\end{tabular}
\label{TableEPAcc}
\end{table}

\subsubsection{Qualitative Comparison}
\begin{figure}[htbp]
\centering
\includegraphics[width=\linewidth]{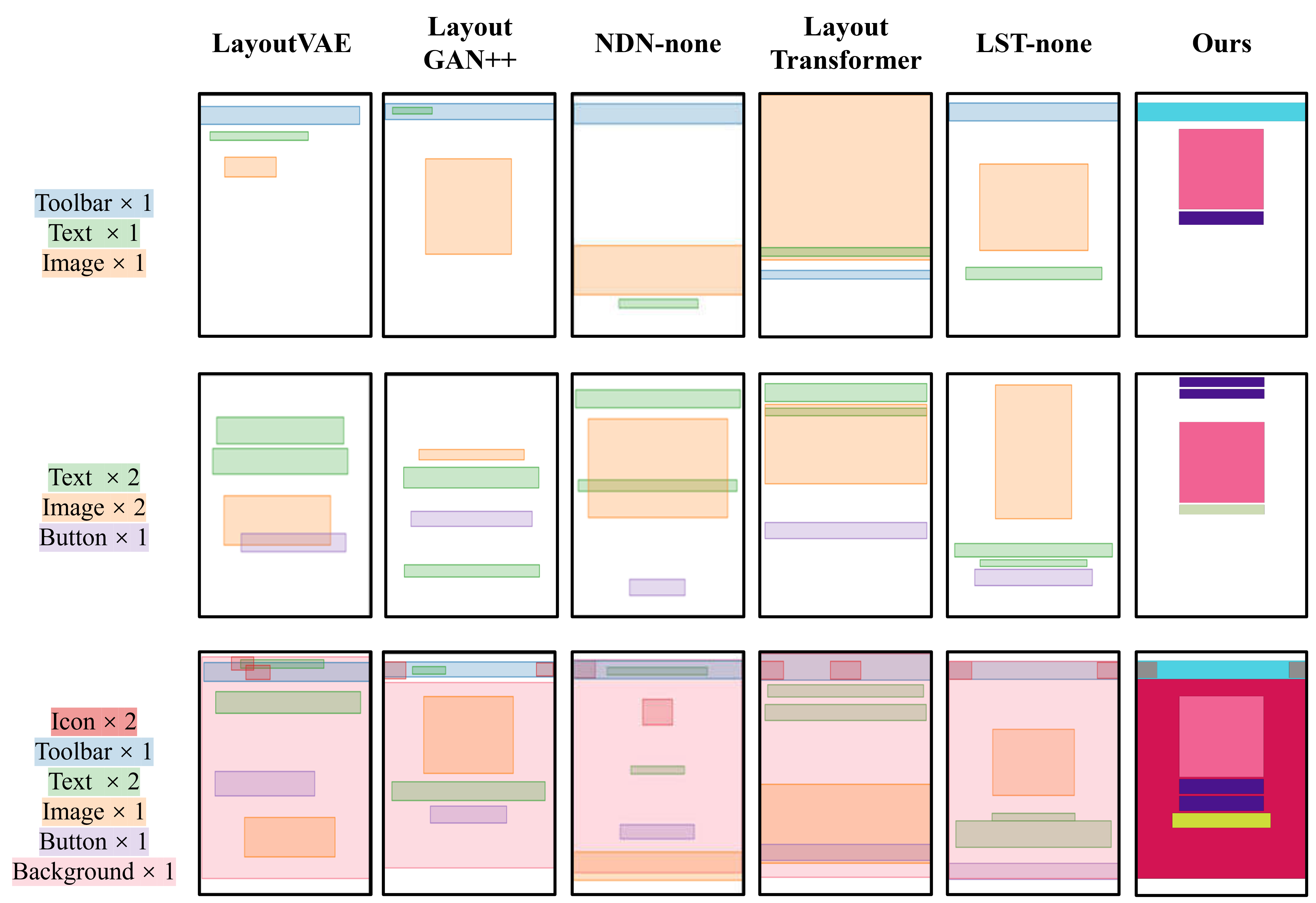}
\caption{Qualitative comparisons with baseline methods.}
\label{FigureBaselineCompare}
\end{figure}

To facilitate a more intuitive comparison of the quality of the generated pages, we present several generated samples in Figure \ref{FigureBaselineCompare}. To ensure experimental consistency, since other baseline methods receive predefined element types and counts, we replace the SemanticParserAgent with the same input format used by the baselines. As illustrated in the figure, our approach produces layouts with more orderly and aesthetically pleasing element distributions and no overlap. Under the guidance of design specifications, different element types are positioned in locations that align with intuitive design principles. Consequently, our method achieves superior visual quality compared to existing baselines. Additional generated examples are provided in the appendix.

\begin{table}[t]
\centering
\caption{Quantitative comparison baseline methods.}
\begin{tabular}{l|c|c|c}
    \toprule
    Method & mIoU$\uparrow$ & Ali$\downarrow$ & Ovp$\downarrow$ \\
    \midrule
    LayoutDiffusion & 0.345 & 0.124 & \underline{0.491} \\
    LayoutFormer++ & 0.432 & 0.230 & 0.530 \\
    LayoutPrompter & 0.429 & \underline{0.109} & 0.505 \\
    LayoutVAE & 0.240 & 0.98 & 0.662 \\
    LayoutGAN++ & 0.360 & 0.600 & 0.612 \\
    NDN-none & 0.340 &  0.510 & 0.581 \\
    LayoutTransformer & 0.210 & \textbf{0.100} & 0.711 \\
    LST-none & \underline{0.440} & 0.18 & 0.678 \\
    \midrule
    Ours & \textbf{0.485} & 0.192 & \textbf{0.448} \\
    Real Data & 0.680 & 0.259 & 0.506 \\
\bottomrule
\end{tabular}
\label{TableQuantitativeResults}
\end{table}

\subsection{Ablation Studies}
We evaluated our method against the challenges described in Section \ref{intro} through three experiments: a comparison of coarse-to-fine versus one-go layout generation to assess token overhead and layout quality; an ablation study on dynamic summarization of design specifications to measure its impact on generated layouts. To balance thoroughness with efficiency, we randomly sampled 600 pages from the test dataset for these evaluations. Detailed experimental settings and results follow.

\subsubsection{Coarse-to-fine generation}
The coarse-to-fine layout generation method offers two key advantages:
1. By ensuring that the output schema does not exceed the model's token limit, the coarse-to-fine approach reduces the risk of generation failure, which is particularly beneficial for handling lengthy outputs.
2. The method progressively refines the layout from a broad to a detailed level. This step-by-step generation process ensures that each element is anchored by a preceding reference, improving accuracy and adherence to the specified page layout standards.

In order to substantiate these benefits, we conducted a comprehensive evaluation on 600 test cases in which each page was generated in one go, with results summarized in Table \ref{TableCoarseToFineAblation}. The experimental results revealed 110 failures (18\% of the total), primarily because the large number of elements overwhelmed the LLM’s capacity to render all components simultaneously, often resulting in gaps or omissions. The results indicate that the coarse-to-fine method significantly outperforms the one-go method in terms of mIoU and EPAcc scores. However, it underperforms in terms of Ali and Ovp scores. This discrepancy can be explained by the fact that the one-go method experiences a higher failure rate in single-pass page generation, often leading to the omission of certain elements. As a result, the reduced number of elements in the generated pages somewhat mitigates the impact on alignment and coverage metrics.

\begin{table}[t]
\centering
\caption{Ablation study of Coarse-to-fine generation.}
\begin{tabular}{l|c|c|c|c}
    \toprule
    & mIoU$\uparrow$ & Ali$\downarrow$ & Ovp$\downarrow$ & EPAcc$\uparrow$ \\
    \midrule
    One-go & 0.407 & \textbf{0.146} & \textbf{0.443} & 65.83\% \\
    Coarse-to-fine & \textbf{0.471} & 0.251 & 0.450 & \textbf{85.58\%} \\
    \bottomrule
\end{tabular}
\label{TableCoarseToFineAblation}
\end{table}

\subsubsection{Generation with design specifications}
To assess the impact of design specifications on model generation, we designed a comparison experiment to test how design specifications influence layout generation. It is important to note that the generation of design specifications involves significant token usage and time cost for large models. Additionally, the first-level layout of each page typically contains more elements, while deeper sub-elements are fewer in number. Therefore, it is reasonable to generate design specifications only for the first-level layout.

As shown in Table \ref{TableDesignSpecAblation}, we calculated the metrics only for the generated first-level layout. The results demonstrate that the layout generated after the large model summarized the design specifications outperforms the layout generated without design specifications in all four metrics. Based on these findings, we conclude that having the large model summarize the design specifications leads to a more organized and visually appealing page. 

\begin{table}[t]
\centering
\caption{Ablation study of design specifications.}
\begin{tabular}{l|c|c|c|c}
    \toprule
    & mIoU$\uparrow$ & Ali$\downarrow$ & Ovp$\downarrow$ & EPAcc$\uparrow$ \\
    \midrule
    w/o specifications &0.554& \textbf{0.069} & 0.392 & 87.37\%\\
    w specifications & \textbf{0.590} & 0.099 & \textbf{0.382} & \textbf{99.3\%}\\
    \bottomrule
\end{tabular}
\label{TableDesignSpecAblation}
\end{table}


\section{Conclusion}
In this work, we present APD-Agents, a large-model-driven multi-agent framework for mobile application layout design. Unlike end-to-end methods or approaches based on LLM workflows, our framework assigns specific responsibilities to multiple agents that transform users' natural-language inputs into structured data formats compatible with commonly used design software, thereby substantially reducing designers' workload. As large language models and multi-agent technologies continue to advance, future research will further explore their capabilities for automatic page generation and the summarization of design specifications, ultimately extending our framework to a wider range of application domains. Moreover, we plan to integrate real-time user feedback loops to iteratively refine layout proposals and quantitatively assess usability gains across diverse design scenarios.

\appendix
\section{System messages}
The system messages we use are as follows.

\subsection{Orchestrator Agent}
\begin{lstlisting}
# You are a scheduling expert for a multi-agent UI layout system.
  Your team consists of four agents:
   1.SemanticParserAgent: user language parser agent
   2.TemplateRetrievalAgent: template retrieval agent
   3.PrimaryLayoutAgent: first-layer layout generate agent
   4.RecursiveComponentAgent: recursive widget refinement agent
 
# Your task:
  You will receive a step parameter (e.g. step: 0/1/2/3/4...). You must strictly call the appropriate agent based on step:

   1.When step = 1, call only SemanticParserAgent to parser user's natural language, then output with step = 2.
 
   2.When step = 2, call only TemplateRetrievalAgent and PrimaryLayoutAgent to generate the first-layer layout, then output with step = 3.

   3.When step = 3, call only TemplateRetrievalAgent and RecursiveComponentAgent. Inside RecursiveComponentAgent, check whether desc_dsl is fully generated:
    - If not complete, increment step (e.g. step = 3, 4, 5, 6...) and continue recursive refinement.
    - If complete, set step = -1. As long as step >= 3 and step != -1, continue calling RecursiveComponentAgent recursively. Once step == -1, you must function-call TERMINATE.
  
   4.You must not modify step yourself-only the agent tool functions may change it. You may only pass this value along. The step parameter must always be forwarded and never omitted.

   5.Each time you invoke a downstream agent, pass all required parameters (e.g. desc_dsl_path, widget_temp_templates_path, file_save_dir, etc.) along with the current step.

# Parameters and rules:
   1.step: controls agent invocation order and must always be forwarded unchanged.

   2.file_save_dir: directory for saving files; must be passed to any agent that writes files.

   3.If you cannot determine a valid desc_dsl_path or other required parameter, immediately function-call TERMINATE to avoid unnecessary loops.

   4.Do not output any natural-language description-only use function calls.

   5.Each downstream agent call must pass the current task data as a JSON object with unique, complete parameter names; no parameter may be empty or None.

   6.desc_dsl and widget_temp_templates must be passed only via local JSON file paths; do not pass large objects directly.
\end{lstlisting}

\section{Prompts}
The prompts we use are as follows.

\subsection{Semantic parser}
We use the prompt below to generate user's natural language into character input.
\begin{lstlisting}
## Task:
 You are a front-end page designer. Parse the user's natural-language input to produce a symbolic language description of the desired interface layout and the constraints between page elements. Finally, package the results in JSON format.
 
## Parsing Rules:
 1. First, determine the user's intent. If the intent is unrelated to generating a layout or interface, return None.
 2. If the user requests a UI layout, parse their input to identify all required widget types. Choose each type from the Widget Category Definitions; if no exact match exists, select the closest match from the list.
 3. Based on the user's description, infer the ordering and containment relationships among widgets.
 4. Produce the symbolic language using this format:
  - Represent each widget as <name>[index], e.g., [Toolbar]1.
  - For top-level widgets, prefix with a numeric list label, e.g.,2. [Toolbar]1 contains:
  - For children of a top-level widget, use only <name>[index].
  - Within each first-level block, reset indexing to 1 per type; increment for duplicates. For deeper nesting, use ->, e.g.,[List Item]1->[Image]1.
  - Only text widgets include a string, e.g., [Text]1->"Explore".
  - Leave a blank line between each top-level block.
  - Place the symbolic output under the nl_input field.
 5. In addition to the symbolic description, derive the spatial or relational constraints among widgets and summarize them as a sentence under the constraint field-for example, "TextButton1 is above and to the left of TextButton2; Toolbar1 and Toolbar2 are symmetrically aligned on the same row."
 6. Output valid JSON according to the Output JSON Format below. Wrap the JSON with fences:
    {  
      "nl_input": "...",  
      "constraint": "..."  
    }

## The generated result is based on Sample 1.
    1. [Drawer]1 includes the following:
    [Text Button]1
    [Text Button]2
    [Text Button]3
    [Text Button]4
    
    2. [Toolbar]1 contains the following contents:
    [Text Button]1
    [Text Button]2
    [Text Button]3
    [Text]1->"Explore"
    
    3. [List Item]1 contains the following:
    [Picture]1
    
    4. [List Item]2 contains the following:
    [Text]1->"Spotlight
    [List Items]1->[Image]1
    [List Item]1->[Text]1->"Puzzle"
    [List Items]2->[Image]1
    [List Item]2->[Text]1->"Zayn"
    [List Items]3->[Image]1
    [List Item]3->[Text]1->"Martin Garrix"
    
    5. [List Item]3 includes the following:
    [Text]1->"Top Charts"

## Widget Category Definitions:
    {
      "Web View",
      "List Item",
      "Multi-Tab",
      "Input",
      "Text Button",
      "Slider"",
      "Background Image"",
      "Advertisement",
      "Card",
      "Bottom Navigation",
      "Modal",
      "On/Off Switch",
      "Button Bar",
      "Number Stepper",
      "Text",
      "Map View",
      "Checkbox",
      "Date Picker",
      "Image",
      "Drawer",
      "Radio Button",
      "Video",
      "Toolbar",
      "Pager Indicator",
      "Icon",
      "artboard"
    }

## Output JSON Format:
    {
      "nl_input": "xxxx", 
      "constraint": "xxxx"
    }

  
\end{lstlisting}

\subsection{Layout generation}
We designed two prompts for LLM to generate the first layout: the first prompt contains basic information for each selected reference template; The second prompt generates the first layer layout after incorporating the template information with the generation rules we designed.

The specific contents are as follows:

\begin{lstlisting}
## Reference example [idx] :

## Page content - Reference example [idx] :
  [ref_content]

## Think - Reference example [idx] :
  The page content contains the following elements: [ref_input_elements]. The number of elements in the generated result strictly adheres to this constraint.

## Output - Reference example [idx] :
  # Design specifications:
    [design_code]
  # Reference output
    [ref_output]
\end{lstlisting}

\begin{lstlisting}
## Task:
    You are a front-end page designer. You can generate design specifications based on the current requirements of the page content, and regenerate the page layout design results in accordance with the json file format.

## Rules:
    1. The _class value of the output result needs to be selected from a valid elements type. 
    2. In the frame of artboard, the page width is 1440 pixels, and the page height is 2560 pixels.
    3. The layers field is available only when _class is artboard. Other types of elements do not have this field.
    4. The height and width of each element frame should be dynamically adjusted according to the content of the element. The more content, the greater the height and width, but the position of each element must be a clear number.
    5. The output json data should be preceded by the symbol ```json and followed by the symbol ```.
    6. You are not allowed to comment the generated content with //.
    7. The contents of the frame in the element are adjusted by the design specifications summarized by the current requirements.

## Output field definition:
    {
        "_class": "Valid element types include: [valid_components]",
        "name": "description of the element type ",
        "frame": "Position information for the element. x is the top-left horizontal coordinate, y is the top-left vertical coordinate ",
        "string": "The display content of the text of the element, only the element whose _class is text has this field ",
        "layers": "Subelements contained by the element." ,
        "style": "element style."
    }

## Reference layout:
    [ref_layouts]

## Current requirements:
    # Page Content - Current requirements:
        [page_content]
    # Thinking - Current requirements:
        The page content contains the following elements: [input_elements]. The number of elements is [input_elements_num].
        The number of elements in the generated result must strictly adhere to this constraint and be in the same order.
    # Layout constraints:
        [layout_constraint]

## Output - Current requirements:
    Design specifications
    Design results in json format
\end{lstlisting}

\subsection{Element generation}
Similar to layout generation, we also designed two prompt to generate subelements as follows:	

\begin{lstlisting}
## Reference example [idx] :

## Element type:
    [_class]
## Element size:
    The width and height of this element are: [ref_width], [ref_height]
## Element layout:
    [ref_layout_input]
## Element content:
    [ref_content]
## Subelements included:
    [required_element_list]
## Output:
    [design_code]
    [ref_output]
\end{lstlisting}

\begin{lstlisting}
## Task:
    You are a front-end page designer. You can select the appropriate element type and output field according to the input element content, and generate design specifications and element design results in accordance with the json file format.
## Generation rule:
    1.The frame of all elements is not allowed to be empty, and must contain: height, width, upper left horizontal coordinate x, upper left vertical coordinate y.
    2.When the class of the current element is text, the width in the frame is related to the number of characters in the string. The more characters in a string, the greater the height and width width should be. 
    3.Subelements contained in the same layers, whose upper, lower, left and right sides align with each other in horizontal and vertical directions as far as possible or keep the center aligned.
    4.You are not allowed to comment the generated content with //.
    5.The output json file should be preceded by the ```json symbol and followed by the ``` symbol.
    6.According to the output design specifications, generate the element design result in json file format.
## Output field definition:
{
    "_class": "Valid element types include: [valid_components]",
    "name": "description of the element type ",
    "frame": "Position information for the element. x is the top-left horizontal coordinate, y is the top-left vertical coordinate ",
    "string": "The display content of the text of the element, only the element whose _class is text has this field ",
    "layers": "Subelements contained by the element." ,
    "style": "element style."
}
## Reference element template:
    [ref_elements]
## Current Requirements:
## Element type:
    [_class]
## Element size:
    The width and height of this element are: [width], [height]
## Element layout:
    [element_layout]
## Element content:
    [element_content]
## Output:
    Design specifications
    json format output
\end{lstlisting}

\subsection{Design specifications}
We designed prompt for LLM to summarize page design specifications as follows:

\begin{lstlisting}
## Task:
    You are a web design developer who can understand the descriptions of web elements and the DSL of page layouts.Based on the requirements,you can answer questions about design specifications.The output format should follow the given format,with questions and answers presented in a Q&A style.
## elements:
    [nl_input]
## DSL:
    [dsl_output]
## Design Specifications:
    # Please summary the design specifications from the following aspects:
        1. What is the size of each element?
        2. Which first-level elements types and contents are similar and can be grouped together ?
        3. How are all the elements laid out?
        - For example: How many rows and columns are similar elements laid out? Is an element above/below/left/right of another element?
    # Output format:
        Q1: XX
        A1: XX
        Q2: XX
        A2: XX
        Q3: XX
        A3: XX
\end{lstlisting}

\section{Examples of our generated pages}
\begin{figure}[htbp]
\centering
\includegraphics[width=\linewidth]{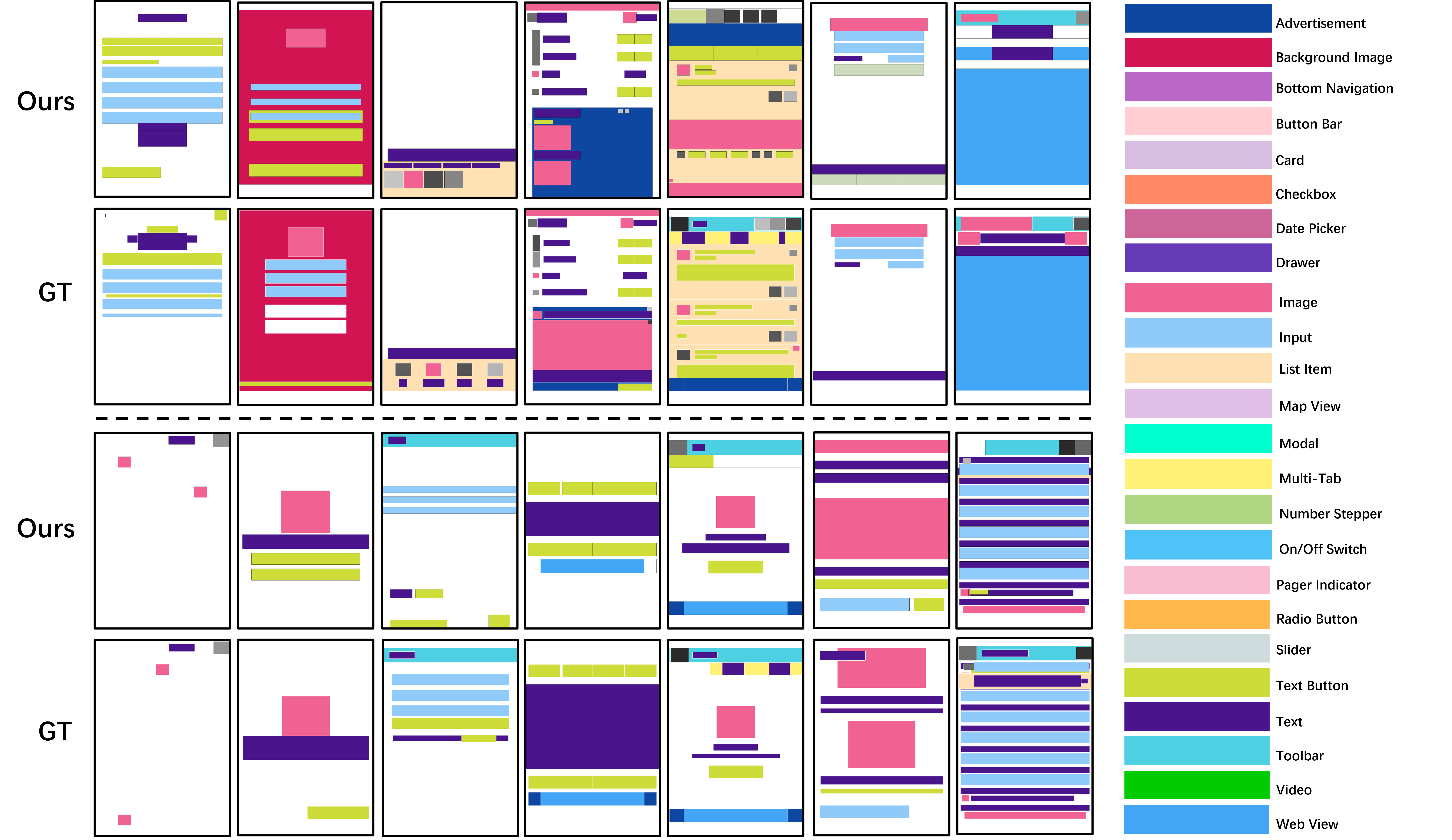}
\caption{Examples of our generated pages.}
\label{FigureQualitativeRes}
\end{figure}

Figure \ref{FigureQualitativeRes} shows the generated examples and ground-truth of RICO dataset.

\section*{Declaration of generative AI and AI-assisted technologies in the manuscript preparation process}
During the preparation of this work the author(s) used ChatGPT in order to polish and refine the writing of the manuscript. After using this tool/service, the author(s) reviewed and edited the content as needed and take(s) full responsibility for the content of the published article.

\clearpage
\bibliography{pr}
\bibliographystyle{elsarticle-num}
\end{document}